\documentclass{article}
\usepackage[preprint,nonatbib]{neurips_2024}

\usepackage{helvet}  % DO NOT CHANGE THIS
\usepackage{courier}  % DO NOT CHANGE THIS
\usepackage[hyphens]{url}  % DO NOT CHANGE THIS
\usepackage{graphicx} % DO NOT CHANGE THIS
\usepackage{caption} % DO NOT CHANGE THIS AND DO NOT ADD ANY OPTIONS TO IT
\usepackage{amsmath, amssymb, amsfonts}
\usepackage{amsthm}
\newtheorem{definition}{Definition}
\newtheorem{theorem}{Theorem}
\newtheorem{lemma}[theorem]{Lemma}
\newtheorem{assumption}{Assumption}

\usepackage{graphicx}
\usepackage{subfigure}
\usepackage{caption}

\usepackage{booktabs, multirow, siunitx, diagbox}

\usepackage{algorithm}
\usepackage{algorithmic}

\usepackage{url}
\title{Set-Valued Sensitivity Analysis of Deep Neural
Networks\thanks{This paper has been accepted to the 39th Annual AAAI Conference on Artificial Intelligence}}

\author{%
  Xin Wang \\
  Department of Civil and Environmental Engineering\\
  University of Washington\\
  Seattle, WA, 98195, United States \\
  {xinw22@uw.edu} \\
  \And
  Feilong Wang \\
  School of Transportation and Logistics\\
  Southwest Jiaotong University\\
  Chengdu, 610032, China \\
  {Flwang@swjtu.edu.cn} \\
  \And
  Xuegang (Jeff) Ban \\
  Department of Civil and Environmental Engineering\\
  University of Washington\\
  Seattle, WA, 98195, United States \\
  {banx@uw.edu} \\
}

\begin{document}

\maketitle

\begin{abstract}
This paper proposes a sensitivity analysis framework based on set-valued mapping for deep neural networks (DNN) to understand and compute how the solutions (model weights) of DNN respond to perturbations in the training data. As a DNN may not exhibit a unique solution (minima) and the algorithm of solving a DNN may lead to different solutions with minor perturbations to input data, we focus on the sensitivity of the solution set of DNN, instead of studying a single solution. In particular, we are interested in the expansion and contraction of the set in response to data perturbations. If the change of solution set can be bounded by the extent of the data perturbation, the model is said to exhibit the Lipschitz-like property. This 'set-to-set' analysis approach provides a deeper understanding of the robustness and reliability of DNNs during training. Our framework incorporates both isolated and non-isolated minima, and critically, does not require the assumption that the Hessian of loss function is non-singular. By developing set-level metrics such as distance between sets, convergence of sets, derivatives of set-valued mapping, and stability across the solution set, we prove that the solution set of the Fully Connected Neural Network holds Lipschitz-like properties. For general neural networks (e.g. Resnet), we introduce a graphical-derivative-based method to estimate the new solution set following data perturbation without retraining.
\end{abstract}

\newpage

\section{Introduction}

% hessian matrix+non-isolated minima-- Fast convergence to non-isolated minima: four equivalent conditions for C2 functions

% morse-bott, psedo inverse-- Non-Convex Bilevel Games with Critical Point Selection Maps

% non-isolated
% GRadient DESCENT METHOd IN NON-CONVEX OPTIMIZATION WITH NON-ISOLATED LOCAL MINIMA*

Sensitivity analysis is a classical research topic that crosses optimization \cite{fiacco1983introduction}, machine learning, and deep learning \cite{yeung2010sensitivity,koh2017understanding,christmann2004robustness}. It studies how the solution of a \textit{model} responds to minor perturbations in hyperparameters or input data. For instance, Christmann et al. \cite{christmann2004robustness} proved that the solution of the classification model with convex risk function is robust to the bias in data distribution. In this paper, we focus on deep neural networks (DNN), for which the solution is the weights of DNN trained (optimized) on input data. We are concerned about how the solution of DNN responds to perturbations in the input data in the \textit{training} stage of the model. 

In the domain of deep learning (e.g., DNN), sensitivity analysis has started to draw attention due to its wide range of applications, such as designing effective data poisoning attacks \cite{munoz2017towards,mei2015using}, evaluating the robustness of models \cite{weng2018evaluating}, and understanding the impact of important features in training data on the prediction \cite{koh2017understanding}. Define a neural network $f: \mathcal{X} \rightrightarrows \mathcal{Y}$, where $\mathcal{X}$ (e.g., images) is the input space and  $\mathcal{Y}$ is the (e.g., labels) output space. Given training data samples $x = [x_1, x_2, \dots, x_n]$ and $y = [y_1, y_2, \dots, y_n]$, and the loss function $L$, the empirical risk minimizer is given by $\hat{w} \stackrel{\text { def }}{=} \arg \min _{w \in \mathcal{W}} \frac{1}{n} \sum_{i=1}^n L\left({x}_i, {y}_i ,w\right)$. This paper assumes that we perturb only the features, keeping the label constant. The learning process from data to local minimizers thus can be formulated as a set-valued mapping $S: \mathcal{X}^n \rightrightarrows \mathcal{W}$ \footnote{Here, $\mathcal{X}^n$ denotes the $n$-fold Cartesian product of $\mathcal{X}$, i.e., $\mathcal{X}^n = \mathcal{X} \times \mathcal{X} \times \cdots \times \mathcal{X}$ ($n$ times).}, 

\begin{equation} 
S({x})=\left\{\hat{w} | \hat{w} \stackrel{\text { def }}{=} \arg \min _{w \in \mathcal{W}} \frac{1}{n} \sum_{i=1}^n L\left({x}_i,{y}_i, w\right)\right\}.
\label{eq:mapping_new}
\end{equation}

For the unperturbed feature $\bar{x} = [\bar{x}_1, \bar{x}_2, \dots, \bar{x}_n]$ and $\bar{w} \in S(\bar{x})$, current sensitivity analysis \cite{koh2017understanding,nickl2024memory,christmann2004robustness} aims to study the change of $S(\bar{x})$ when an individual point is perturbed. For example, if the feature $\bar{x}$ is perturbed to $\hat{x}=[\bar{x}_1,\bar{x}_2,\dots,x_p,\dots,\bar{x}_n]$, the sensitivity analysis can be conducted by examining a limit:
\begin{equation}\label{limit}
    \lim _{x_p \rightarrow \bar{x}_p} \frac{S\left(\hat{x}\right)-S(\bar{x})}{\|x_p-\bar{x}_p\|}.
\end{equation}
Most current sensitivity analysis methods for DNN suffer from the following two issues. First, to figure out the sensitivity of solution ${w}$ w.r.t. data ${x}_p$, one of the most common approaches (e.g., the influence function \cite{koh2017understanding}) is to apply the Dini implicit function theorem \cite{dontchev2009implicit} to the optimality condition of the model: $\frac{1}{n} \sum_{i=1}^n \nabla_w L\left(\bar{x}_i, \bar{y}_i, \bar{w}\right)=0$, which leads to:
\begin{equation}\label{Dini}
\begin{aligned}
 \lim _{x_p \rightarrow \bar{x}_p} \frac{S\left(\hat{x}\right)-S(\bar{x})}{\|x_p-\bar{x}_p\|} &= \nabla_{x_p} w \bigg|_{x_p=\bar{x}_p} \\
&=-H_{\bar{w}}^{-1} \nabla_{x_p} \nabla_\theta L(\bar{x}_p,\bar{y}_p, \bar{w}),
\end{aligned}
\end{equation}
where $H_{\bar{w}}^{-1}=\frac{1}{n} \sum_{i=1}^n \nabla_w^2 L\left(\bar{x}_i,\bar{y}_i, \bar{w}\right)$ is the Hessian.
However, the application of the Dini implicit function theorem is invalid when the Hessian $H_{\bar{w}}$ is not invertible due to the often non-locally strong convex loss function $L$ of DNN \cite{li2018visualizing}. Second, the current approaches \cite{koh2017understanding,nickl2024memory} assume $S$ is a single-valued mapping, omitting the fact that  $S$ is often set-valued. Some findings have demonstrated that DNN may not exhibit a unique solution $S(x)$ (even when $L$ is convex). For example, it was noticed in \cite{li2018visualizing} that the stochastic gradient descent (SGD) method can find flat minima (solutions located in the flat valley of the loss landscape); others found that all SGD solutions for DNN may form a manifold \cite{benton2021loss,cooper2021global}. When an application (e.g., a data poisoning attack) is designed and evaluated based on only one of the solutions, it overlooks the fact that the learning algorithm may converge to other solutions during re-training. 

In this paper, we incorporate the fact that $S$ is often set-valued into the sensitivity analysis framework for DNNs. This extends the scope of sensitivity analysis in DNNs from focusing on a single solution to a solution set, shifting from the traditional 'point-to-point' analysis to a 'set-to-set' paradigm. That is, we study how the solution set of a DNN expands and contracts in response to data perturbations. The proposed approach covers more general situations in risk minimization of DNN, including isolated local minima, non-isolated minima, and minima that consist of a connected manifold. More importantly, it directly deals with the solution sets without the assumption of non-singular Hessian matrix, offering a more complete understanding of DNN.

% Our 'set-to-set' paradigm considers the overall shift in the dataset, allowing multiple points to be perturbed simultaneously. In this case, a natural idea is to utilize the limit (\ref{limit}) to replace the limit (\ref{limit}) when data is perturbed from $\bar{x}$ to $x$. \footnote{Here, $\left\|x - \bar{x}\right\|$ represents the square root of the sum of the squares of the differences between corresponding elements, i.e., $\left\|x - \bar{x}\right\| = \sqrt{\sum_{i=1}^{n} \|x_i - \bar{x}_i\|^2}$.}:

% \begin{equation}\label{limit_2}
%     \lim _{x \rightarrow \bar{x}} \frac{S({x})-S(\bar{x})}{\left\|x-\bar{x}\right\|}
% \end{equation}

Considering $S$ as set-valued, with the isolated minima a special case where solution set $S(x)$ contains only one element, the sensitivity analysis aims to study 1) whether the limit (\ref{limit}) exists, and 2) whether the limit (\ref{limit}) is bounded. However, directly studying the limit when mapping $S$ is set-valued could be challenging. On the one hand, $S(x)$ could be a large or even infinitely large set (e.g., a manifold), which makes it complicated to analyze. To address this, we instead focus on the change of the solution set $S(x)$ within a small neighborhood. Given a pair $(\bar{x}, \bar{w}), \bar{w} \in S(\bar{x})$, we study the change of $S(x) \cap U$ , where $U$ is a neighborhood of $\bar{w}$, in response to the data perturbation within a neighborhood $V$ of $\bar{x}$. On the other hand, as $S(x)$ and $S(\bar{x})$ are sets rather than single solutions, the subtraction operation between sets does not exist (set operations only include union, intersection, and set difference), which invalidates the limit (\ref{limit}). Consequently, we leverage the \textit{Lipchitz-like} property of $S$ around $(\bar{x},\bar{w})$, to measure the change of $S(x)\cap U$. We say that $S$ holds \textit{Lipchitz-like} property around $\left(\bar{x}, \bar{w}\right) \in \operatorname{gph} S:=\{(x, w) \mid w \in S(x)\}$, if there exist neighborhoods $U$ of $\bar{w}, V$ of $\bar{x}$ and a positive real number $\kappa$ such that \cite{dontchev2009implicit} \footnote{Here, $\left\|x - {x^\prime}\right\|$ represents the square root of the sum of the squares of the differences between corresponding elements, i.e., $\left\|x - {x^\prime}\right\| = \sqrt{\sum_{i=1}^{n} \|x_i - {x^\prime}_i\|^2}$.}
\begin{equation}\label{def:aubin}
S(x^\prime) \cap U \subset S(x)+\kappa\left\|x-x^{\prime}\right\| \mathbb{B}, \quad \forall x^{\prime}, x \in V,
\end{equation}
where $\mathbb{B}$ is a closed unit ball. The Lipschitz-like property is an extension of Lipschitz continuity defined for single-valued functions to set-valued mappings. $\kappa$, as a scalar, is the \textit{Lipschitz module} that describes the upper bound of the solution set's change in response to data perturbations. The existence of the limit (\ref{limit}) can be interpreted as the establishment of a Lipschitz-like property, and the bound of the limit (\ref{limit}) is equivalent to bounded $\kappa$. Note also that $\kappa$ characterizes how sensitive the model solution is w.r.t input data, which is defined for the training stage. It complements the Lipschitz constants of DNN studied in \cite{fazlyab2019efficient,virmaux2018lipschitz} that have been only defined and studied so far for the inference stage (with fixed model solution).

In our "set-to-set" analysis framework, the study of existence and boundedness of limits (\ref{limit}) is transferred as the following question: Given a local minimizer $\bar{w}$ with its neighborhoods $\mathrm{U}$, and training data $\bar{x}$ with its neighborhoods $\mathrm{V}$, where $(\bar{x},\bar{w}) \in  \operatorname{gph} S$, does the solution mapping $S$ satisfy the Lipschitz-like property in a neighborhood of $(\bar{x}, \bar{w})$, and if so, how can we estimate the associated bounded Lipschitz modulus? Moreover, we also explore how, when we perturb the data $\bar{x}$ to $x^p$, with $x^p\in V$, we can identify the new solution set $S\left(x^p\right) \cap U$ without re-training.

We prove that, for the Deep Fully Connected Neural Network (DFCNN) with the Relu activation function, the solution set $S$ holds the Lipschitz-like property. A bound of the Lipschitz module is also provided. This reveals that the solution set of a DFCNN will not deviate significantly when there are biases in the training data, allowing us to estimate the new solution set based on the behavior of $S$ around $(\bar{x},\bar{w})$. We also introduce \textit{graphical derivatives} to capture the local linear approximation of $S$ around $(\bar{x},\bar{w})$ for a DNN (not only for DFCNN). The graphical derivative based method can accurately estimate $S(x^p)-\bar{w}$, providing solutions following data perturbations for a DNN (e.g., Resnet56) with near zero training loss. In particular, when $S(x^p)$ is single-valued, i.e. the solution set $S(x^p)$ within neighborhood $U$ includes only one solution, our graphical derivative based method is equivalent to the influence function \cite{koh2017understanding}.

The contribution of our paper is summarized as follows:

1) We introduce a set-valued mapping approach to understand the sensitivity of solutions of DNN in relation to perturbations in the training data. Our framework accommodates both isolated and non-isolated minima without relying on convex loss assumption. 

2) We prove that the solution mapping of DFCNN holds the Lipschitz-like property and estimate a bound for the Lipschitz module. This implies that DFCNN is stable during training.

3) We propose a graphical-derivative-based method to estimate the new solution set of a DNN when the training data are perturbed, and simulate it using the Resnet56 with the CIFAR-10 dataset.

\section{Preliminary knowledge}\label{sec:pre}

This section briefly summarizes the necessary preliminary knowledge for sensitive analysis of set-valued mapping, covering the distance between sets, sets convergence, and the generalized derivative. %The generalized derivative includes the graphical derivative and the coderivative. 
These concepts will used to characterize the behavior of set-valued mapping and play a key role in defining the criteria for Lipschitz-like property. 

% \textbf{Notation}: For a neural network $f: X \rightrightarrows Y$, where $\mathcal{X}$ (e.g., images) is the input space and  $\mathcal{Y}$ is the(e.g., labels) output space. Given training points $\left[\left(x_1, y_1\right),\left(x_2, y_2\right) \ldots\left(x_n, y_n\right)\right]$ and loss function $L$, the empirical risk minimizer is given by $\bar{w} \stackrel{\text { def }}{=} \arg \min _{w \in \mathcal{W}} \frac{1}{n} \sum_{i=1}^n L\left(z_i, \theta\right)$. Since only $x=[x_1,\dots,x_n]$ will be perturbed in this paper(label is constant), the set-valued mapping defined by (\ref{eq:mapping}) is equivalent to:
% \begin{equation}
% S(x)=\{\bar{w} | \bar{w} \stackrel{\text { def }}{=} \arg \min _{w \in \mathcal{W}} \frac{1}{n} \sum_{i=1}^n L\left(x_i,y_i, w\right)\}
% \label{eq:mapping_new}
% \end{equation}

\begin{definition}\textup{\textbf{(Distance between sets)}}\label{Def1}
The distance from a point $w$ to a set $C$ is
\begin{equation*}
    d_C(w)=d(w, C)=\inf _{w^\prime \in C}|w-w^\prime|.
\end{equation*}
For sets $C$ and $D$, the Pompeiu-Hausdorff Distance of $C$ beyond $D$ is defined by
\begin{equation*}
    h(C, D)=\max \{e(C, D), e(D, C)\}, e(C, D)=\sup _{w \in C} d(w, D),e(D, C)=\sup _{w \in D} d(w, C).
\end{equation*}
\end{definition} 

% \textbf{Example 1:}Consider two sets $C$ and $D$ on a number line: $C=\{1,2,3\}$ and $D=\{4,5\}$. The distance from $C$ to $D$ would be:
% $$
% e(C, D)=\sup _{w \in C} d(w, D)=\max \{|1-4|,|2-4|,|3-4|\}=\max \{3,2,1\}=3
% $$

% Similarly, the distance from $D$ to $C$ would be:
% $$
% e(D, C)=\sup _{w \in D} d(w, C)=\max \{|4-3|,|5-3|\}=\max \{1,2\}=2
% $$

\begin{figure}
    \centering
    \includegraphics[width=0.4\linewidth]{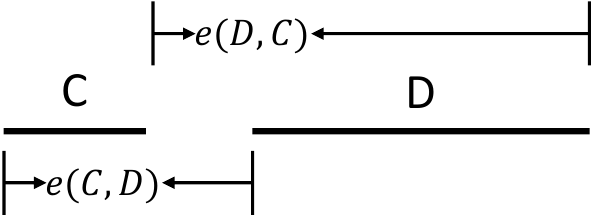}
    \caption{Illustration of $e(C,D)$ and $e(D,C)$, where $C$ and $D$ are two closed sets. The Pompeiu-Hausdorff Distance of $C$ beyond $D$ is $h(C, D)=\max \{e(C, D), e(D, C)\}=e(D,C)$.}
    \label{fig:def1}
\end{figure}

% Thus, the Pompeiu-Hausdorff distance between $C$ and $D$ is:
% $$
% h(C, D)=\max \{e(C, D), e(D, C)\}=\max \{3,2\}=3
% $$
An illustrative figure is provided in Figure \ref{fig:def1}, where $C$ and $D$ are two segments.
\begin{definition}\textup{\textbf{(Painlevé-Kuratowski Set convergence)}}
Given a set-valued mapping $S: X \rightrightarrows W$, the Painlevé-Kuratowski outer set limit as $x \rightarrow \bar{x} $ is
\begin{equation*}
\begin{aligned}
\limsup_{x \to \bar{x}} S(x):=\{w \in W \mid & \exists \text { sequences } x_k \rightarrow \bar{x} \text { s.t. } w_k \in S\left(x_k\right)\rightarrow w \},
\end{aligned}
\end{equation*}

the Painlevé-Kuratowski inner set limit as $x \rightarrow \bar{x} $ is
\begin{equation*}.
\begin{aligned}
\liminf_{x \to \bar{x}} S(x):=\{w \in W \mid & \text{ for }  \forall \text { sequences } x_k \rightarrow \bar{x}, w_k \in S\left(x_k\right)  \rightarrow w\}.
\end{aligned}
\end{equation*}
\end{definition}

\begin{definition}\label{def3}
   A vector $\eta$ is tangent to a set $\Gamma$ at a point $\bar{\gamma} \in \Gamma$, written $\eta \in T_\Gamma(\bar{\gamma})$, if
\begin{equation*}
\frac{\gamma_i-\bar{\gamma}}{\tau_i} \rightarrow \eta \text { for some } \gamma_i \rightarrow \bar{\gamma}, c_i \in \Gamma, \tau_i \searrow 0 \text {. }
\end{equation*}
\end{definition}
Where $T_\Gamma(\bar{\gamma})$ is the tangent cone to $\Gamma$ at $\bar{\gamma}$. 
\begin{definition}\label{def4}
    Given a convex set $\Gamma$ in $\mathbb{R}^n$ and a point $\bar{\gamma}$ in $\Gamma$, the normal cone to $\Gamma$ at $\bar{\gamma}$, denoted $N_\Gamma(\bar{\gamma})$, is defined as the set of all vectors $\xi \in \mathbb{R}^n$ that satisfy the condition:
\begin{equation*}
N_\Gamma(\bar{\gamma})=\left\{\xi \in \mathbb{R}^n:\langle \xi, \gamma-\bar{\gamma}\rangle \leq 0 \text { for all } \gamma \in \Gamma\right\}.
\end{equation*}
\end{definition}

\textbf{Remark 1:} This study only focuses on the normal cone and tangent cone of convex sets. Refer to Appendix \ref{appendix C} for an illustrative diagram of Definition 3 and Definition 4.

We next define the generalized derivative that includes both the graphical derivative and coderative. The graphical derivative is a concept used in the analysis of set-valued mappings (multifunctions). It generalizes the derivative of functions to set-valued mappings, which captures the local behavior of a set-valued mapping around a particular point. The coderivative complements the graphical derivative, providing information about how small changes in the output can affect the input.

\begin{definition}\textup{\textbf{(Generalized derivatives)}}\label{def:genedev}\footnote{$A^*$ denotes the conjugate transpose of $A$, where $A$ is a matrix}
% Given a set-valued mapping $S: \mathbb{R}^n \rightrightarrows \mathbb{R}^m$ and a point $(\bar{x}, \bar{y}) \in \operatorname{gph} F$ from its graph, the graphical derivative of $S$ at $\bar{x}$ for $\bar{y}$ is the mappings $DF(\bar{x} \mid \bar{y}): \mathbb{R}^n \rightrightarrows \mathbb{R}^m$ with the values
% \begin{equation*}
%  DS(\bar{x} \mid \bar{y})(v):=\limsup_{\substack{h \rightarrow v \\ \tau \downarrow 0}} \frac{S(\bar{x}+\tau h)-\bar{y}}{\tau} .   
% \end{equation*}
% The coderivative at $(\bar{x}, \bar{y}) \in \operatorname{gph}(S)$ is denoted by $D^* S(\bar{x} \mid \bar{y}): \mathbb{R}^m \rightrightarrows \mathbb{R}^n$ and is defined by
% $$
% v \in D^* S(\bar{x} \mid \bar{y})(u) \Leftrightarrow(v,-u) \in N_{\operatorname{gph}(S)}(\bar{x}, \bar{y}) .
% $$
\label{def:derivative} 
Consider a mapping $S: \mathbb{R}^n \rightrightarrows \mathbb{R}^m$ and a point $\bar{x} \in \operatorname{dom} S$.  The \textbf{graphical derivative} of $S$ at $\bar{x}$ for any $\bar{w} \in S(\bar{x})$ is the mapping $D S(\bar{x} \mid \bar{u}): \mathbb{R}^n \rightrightarrows \mathbb{R}^m$ defined by
\begin{equation*}
    v \in D S(\bar{x} \mid \bar{w})(\mu) \Longleftrightarrow(\mu, v) \in T_{\operatorname{gph} S}(\bar{x}, \bar{w}),
\end{equation*}

whereas the \textbf{coderivative} is the mapping $D^* S(\bar{x} \mid \bar{u}): \mathbb{R}^m \rightrightarrows \mathbb{R}^n$ defined by
\begin{equation*}
    q \in D^* S(\bar{x} \mid \bar{w})(p) \Longleftrightarrow(q,-p) \in N_{\operatorname{gph} S}(\bar{x}, \bar{w}) .
\end{equation*}

\textbf{Remark 1:} Graphical derivative can also be expressed as: 

\begin{equation}
    DS(\bar{x}\mid\bar{w})(\mu)=\limsup_{\substack{\tau \searrow 0 \\ \mu_k \rightarrow \mu}} \frac{S(\bar{x}+\tau \mu_k)-\bar{w}}{\tau}
\end{equation}

\textbf{Remark 2:} In the case of a smooth, single-valued mapping $F: \mathbb{R}^n \rightarrow \mathbb{R}^m$, one has
\begin{equation}
\begin{aligned}
& D F(\bar{x})(\mu)=\nabla F(\bar{x}) \mu \text { for all } \mu \in \mathbb{R}^n \\
& D^* F(\bar{x})(p)=\nabla F(\bar{x})^* p \text { for all } p \in \mathbb{R}^m
\end{aligned}
\end{equation}

\end{definition}

\section{Lipschitz-like property of Deep Fully Connected Neural Network}
\label{sec:MLP}
This section studies the Lipschitz-like property of DNNs. We focus on DFCNN, a classical DNN, to demonstrate our main theorem results. For each data point $x_{i} \in x, x_i \in \mathbb{R}^d $, the first layer's weight matrix of a DFCNN is denoted as ${W}^{(1)} \in \mathbb{R}^{m \times d}$, and for each subsequent layer from $2$ to $H$, the weight matrices are denoted as ${W}^{(h)} \in \mathbb{R}^{m \times m}$. $a \in \mathbb{R}^m$ is the output layer and the Relu function is given by $\sigma(\cdot)$. We recursively define a DFCNN, starting with ${x_i}^{(0)}={x_i}$ for simplicity.
\begin{equation}
\begin{aligned}
{x_i}^{(h)} & =\sigma\left({W}^{(h)} {x_i}^{(h-1)}\right), 1 \leq h \leq H \\
f({x_i}, w) & ={a}^{\top}{x_i}^{(H)} .
\end{aligned}
\end{equation}
Here ${x_i}^{(h)}$ is the output of the $h$-th layer. We denote $W:=(W^{(1)},\dots,{W}^{(H)})$  as the weights of the network and $w :=(w^{(1)},\dots,{w}^{(H)})$ as the vector of the flatten weights. In particular, $w^{(h)}$ is the vector of the flattened $h$-th weight $W^{(h)}$. Denote $dim(w^{(h)})=p^{(h)}$ and  $dim(w)=\sum_{i=1}^Hp^{(h)}=p$. 

For a DFCNN, we develop our method using the quadratic loss function: $L(x_i,y_i,w)=\frac{1}{2}\left(f\left(w, {x}_i\right)-y_i\right)^2$. $w$, as the neural network weights, is a local/global minimum of loss $L$. Since first-order optimization algorithms, such as SGD, are widely utilized, we employ the first-order optimality condition to characterize these minima. Let $R(x,y,w)=\nabla_w \frac{1}{n} \sum_{i=1}^n L\left(x_i,y_i, w\right)$. Since th label vector $y = [y_1, \dots, y_n]$ is constant, we simplify the notation of $R(x, y, w)$ to $R(x, w)$. Then the solution of a DFCNN can be characterized by the set-valued mapping $F$ below:
\begin{equation}
\begin{aligned}
F(x)&=\{{w} |R(x, {w}) = \nabla_w \frac{1}{n} \sum_{i=1}^n L\left(x_i,y_i, w\right)=0\},
\end{aligned}
\end{equation}
For layer $h$, we define mapping $F_h$ as:
\begin{equation}
F_h(x)=\{w^{(h)} | R(x,{w})=0\}.
\end{equation}

Following the sensitivity analysis in \cite{koh2017understanding,nickl2024memory}, we first focus on one individual data $x_k \in x=[x_1,\dots,x_n]$, which is perturbed. In this case, $F(x)$ and $F_h(x)$ in the above two equations are expressed as $F(x_k)$ and $F_h(x_k)$ to indicate that only $x_k$ are perturbed. In Section \ref{sec:4}, we present the case with multiple data perturbations.

\begin{assumption}
We assume that DNNs are overparameterized; under this assumption, a DFCNN has the capacity to memorize training data with zero training error, i.e. $p > d$. 
\end{assumption}

\begin{assumption}
   For given $\bar{x}$ and $\bar{w}$, $[\nabla_w R(\bar{x},\bar{w}),\nabla_{x_k}R(\bar{x},\bar{w})]$ is of full rank, where $\nabla_w R(\bar{x},\bar{w}) \in \mathbb{R}^{p\times p}$, $\nabla_{x_k}R(\bar{x},\bar{w}) \in \mathbb{R}^{p\times d} , p=dim(w), d=dim(x_k)$.
\end{assumption} 
\textbf{Remark 1:} It can be demonstrated that a DNN with non-singular Hessian matrix, as a special case, fulfills Assumption 2.

\textbf{Remark 2:} In Section \ref{sec:4}, we will find that although some DNNs do not always satisfy Assumptions 1 or 2, our algorithm can still provide a reasonably accurate estimation of the sensitivity of the solution mapping.

The Lipschitz-like property examination and solution set estimation following data perturbation rely on the generalized derivative (see definition \ref{def:derivative}). Theorem \ref{themrem:1} below provides an explicit formulation for the generalized derivative of $F$, enabling a convenient analysis of the local behavior of the solution mapping. It will be used in the proof of Theorem \ref{thm:2} that describes the Lipschitz-like property of the given solution mapping.

\begin{theorem}\label{themrem:1}
    For given $\bar{x}_k$ and $\bar{w}$, the graphical derivative and coderivative of F (defined by Definition \ref{def:derivative}) at $\bar{x}_k$ for $\bar{w}$ have the formulas:

\begin{equation}
\begin{aligned}
&D F\left(\bar{x}_{k} \mid \bar{w}\right)(\mu)=\left\{v \mid \nabla_{w} R(\bar{x}, \bar{w}) v+\nabla_{x_k} R(\bar{x}, \bar{w}) \mu=0\right\}\\
&D^* F\left(\bar{x}_{k} \mid \bar{w}\right)(p)=\left\{\nabla_{x_k}R(\bar{x},\bar{w})^{\top}y \mid y \in \mathbb{R}^{dim(w)}, p+\nabla_{w} R(\bar{x}, \bar{w})^{\top} y=0\right\}
\end{aligned}
\end{equation}

Proof: see appendix \ref{appendix A}.
\end{theorem}

The following Theorem \ref{thm:2} then proves the Lipschitz-like property of DFCNN with the bound of the Lipshitz module. It reveals the potential training stability of DFCNN, i.e. the solution set will not change dramatically when perturbations are introduced to training data. The Lipschitz module is determined by the original solution $\bar{w}$ and input data $\bar{x}$. %Note that we are only concerned with the weights of one layer, e.g. $w_h$, to convenience our discussion.
\begin{theorem}\label{thm:2}
    For a given layer $h$ ($1 \le h \le H$), mapping $F_h$ holds the Lipschitz-like property. For given $\bar{x}_k$ and $\bar{w}$, there exists neighborhoods $V_{k}$ of $\bar{x}_k$ and $U_h$ of $\bar{w}^{(h)}$, with a positive real number $\kappa_h$ (i.e., the Lipschitz modulus) such that

\begin{equation}\label{eq:thmmlp}
F_h\left(x_k^{\prime}\right) \cap U_h \subset F_h(x_k)+\kappa_h\left\|x_k-x_k^{\prime}\right\| \mathbb{B} \quad \forall x_k^{\prime}, x_k \in V_{k}
\end{equation}
where the the Lipschitz modulus $\kappa_h$ can be expressed as:
\begin{equation*}
\begin{aligned}
\kappa_h & =\frac{\left\|\left[\prod_{k=1}^H W^{(k)^{\top}} \operatorname{diag}\left(\mathbf{1}\left(\sigma\left(W^{(k)} x_k^{k-1}\right)>0\right)\right)\right] a\right\|}{\| \operatorname{diag}\left(\mathbf{1}\left(\sigma\left(W^{(h)} x_k^{h-1}\right)>0\right)\right)} \\
& {\left[\prod_{k=h+1}^H W^{(k)^{\top}} \operatorname{diag}\left(\mathbf{1}\left(\sigma\left(W^{(k)} x_k^{k-1}\right)>0\right)\right)\right] a x_k^{(h-1)^{\top}} \|_F }
\end{aligned}
\end{equation*}
% \begin{equation*}
% \kappa_h & =\frac{\left\|\left[\prod_{k=1}^H W^{(k)^{\top}} \operatorname{diag}\left(\mathbf{1}\left(\sigma\left(W^{(k)} x_k^{k-1}\right)>0\right)\right)\right] a\right\|}{\left\|\operatorname{diag}\left(\mathbf{1}\left(\sigma\left(W^{(h)} x_k^{h-1}\right)>0\right)\right)\left[\prod_{k=h+1}^H W^{(k)^{\top}} \operatorname{diag}\left(\mathbf{1}\left(\sigma\left(W^{(k)} x_k^{k-1}\right)>0\right)\right)\right] a x_k^{(h-1)^{\top}}\right\|_F}
% \end{equation*}
Proof: see appendix \ref{appendix B}.

\end{theorem}

\begin{theorem}\label{thm3}
For given $\bar{x}$ and $\bar{w}$, there exists neighborhoods $V$ of $\bar{x}$ and $U$ of $\bar{w}^{(h)}$, with a positive real number $\kappa$ (i.e., the Lipschitz modulus) such that
\begin{equation}
F_h\left(x^{\prime}\right) \cap U \subset F_h\left(x\right)+\kappa\left\|x-x^{\prime}\right\| \mathbb{B} \quad \forall x^{\prime}, x \in V
\end{equation}
Proof: see appendix \ref{appendix B}.
\end{theorem}

% [\textbf{JB}: some discussions on the theorem: further explanations, its physical meaning, practical implications, etc.]

\section{Sensitivity analysis for solution set}\label{sec:4}

Theorem \ref{thm3} reveals that the solution set of DFCNN after perturbation will not deviate dramatically from the original solution set, allowing us to approximate this change using the local information around $(\bar{x},\bar{w})$. This section first proposes a method to estimate the new solution set of DFCNN, given the perturbation in the training data. Unlike Section \ref{sec:MLP}, which only perturbs a single individual data point, this section perturbs multiple data points simultaneously.

Consider a DFCNN with its weight $\bar{w}$, where $\bar{w}$ represents a solution obtained by a learning algorithm (e.g., SGD) trained on a set of pristine data $\bar{x} = \{\bar{x}_i\}, i \in I = \{1, 2, \dots, n\}$.  Assume the data is perturbed following $x_i^p=\bar{x}_i+\delta \Delta x_i$, where $\delta$ is the norm of perturbation, $\Delta x_i$ is a unit vector that indicates the perturbation direction for point $x_i$. We denote by $\Delta x=\{\Delta x_i\}, i\in I$ the set of perturbations. To make the number of perturbed points more flexible, we set $K \subset I$ to denote the indices of the perturbed data, and let $\Delta x_i=0$ for $i \in I \setminus K$. As defined by (\ref{eq:mapping_new}), $S(x^p)$ is the solution set of a DFCNN trained by the poisoned data $x^p=\{x_i^p\}$ and $S(\bar{x})$ is the original solution set.

The graphical derivative $DS(\bar{x}\mid \bar{w})(\mu)$ captures how the solution $w$ changes near $\bar{w}$ when $x$ is perturbed in the direction of $\mu$. For any twice differentiable loss function $L(w)$ (such as the quadratic loss function), following theorem \ref{themrem:1}, $DS(\bar{x}\mid \bar{w})(\mu)$ is equivalent to (see Appendix \ref{appendix A}):
\begin{equation}\label{eq:solution estimation}
    DS(\bar{x}\mid \bar{w})(\mu)=\left\{v \mid\ \nabla_w^2\frac{1}{|I|} \sum_{i\in I} L\left(x_i, y_i, w\right)v+\nabla_x\nabla_w\frac{1}{|K|} \sum_{i\in K} L\left(x_i, y_i, w\right)\mu=0\right\}.
\end{equation}
The solution set $S(x^p)$ can be estimated by:
\begin{equation}\label{solution estimated method}
    S(x^p)\approx \bar{w}+DS(\bar{x}\mid \bar{w})(\Delta x).
\end{equation}
In a special case, when the empirical risk function $\sum_{i=1}^n L\left(x_i, y_i, w\right)$ has a non-singular Hessian matrix, $D S(\bar{x} \mid \bar{w})(\mu)$ will include a unique $v$, indicating that the solution $\bar{w}$ can only move along direction $v$ if we perturb $\bar{x}$ along direction $\mu$. This situation corresponds to the scenario where the solution $\bar{w}$ represents an isolated minimum. In this case, multiplying both sides of expression on the right side of (\ref{eq:solution estimation}) by the inverse of the Hessian matrix results in the influence function \cite{koh2017understanding}.

\begin{algorithm}[H]
\caption{Training Stability Analysis Based on Set Valued Mapping}
{\textbf{Input:} Perturbation index set $K$, data perturbation $\Delta x$, original solution $\bar{w}$, unperturbed data $\bar{x}$, loss function $L$}
\label{alg}

{\textbf{Output:} Distance between the new solution set and $\bar{w}$}

\begin{enumerate}
    \item Compute the graphical derivative $D S(\bar{x} \mid \bar{w})(\Delta x)$ following (\ref{eq:solution estimation}).

    \item  Estimate the new solution set by $S\left(x^p\right) \approx \bar{w}+D S(\bar{x} \mid \bar{w})(\Delta x)$ 

    \item {Calculate distance} between the solution set and $\bar{w}$ following Pompeiu-Hausdorff Distance (see definition \ref{Def1})
\end{enumerate}

\end{algorithm}

\section{Simulation for solution estimation}
Although our theoretical results in Section \ref{sec:MLP} and Section \ref{sec:4} focus on DFCNN, this section demonstrates that the methods perform well for general DNNs. To show this, we next present two numerical examples to illustrate the proposed set-valued sensitivity analysis method. The first one is on a toy example and the second one is on the Resnet. All experiments are performed on an RTX 4090 GPU.

\subsection{A toy example}

We consider a linear neural network with 2 layers. Assume we only have two data points $\left(x_i, y_i\right), i=1,2$ and both $x_i$ and $y_i$ are real numbers. The solution set $S(x)=\{w=\left(w_1, w_2\right)\}$, where $w_1,w_2$ are the weights of the first and second layer, is given by minimizing the empirical risk :
\begin{equation}
    \begin{aligned}
& \left(w_1, w_2\right) \triangleq \underset{w_1, w_2 \in R}{\operatorname{argmin}} \frac{1}{2}\left(y_1-w_1 w_2 x_1\right)^2+\frac{1}{2}\left(y_2-w_1 w_2 x_2\right)^2.
\end{aligned}
\end{equation}
Given the pristine dataset $\left(\bar{x}_1, \bar{y}_1\right)=(1,2),\left(\bar{x}_2, \bar{y}_2\right)=(2,4)$, the model solution constitutes a set as $w_1*w_2=2$, and $\bar{w}=(1,2)$ is obviously one of the solutions. We assume that the original solution converges to $\bar{w}=(1,2)$ during training using this pristine data. We introduce perturbations to the data following the rule $x^p=\bar{x}+0.2*(-1,-2)$, and re-train the model using the poisoned data.

\begin{figure}[htbp]
    \centering
    \includegraphics[width=0.6\linewidth]{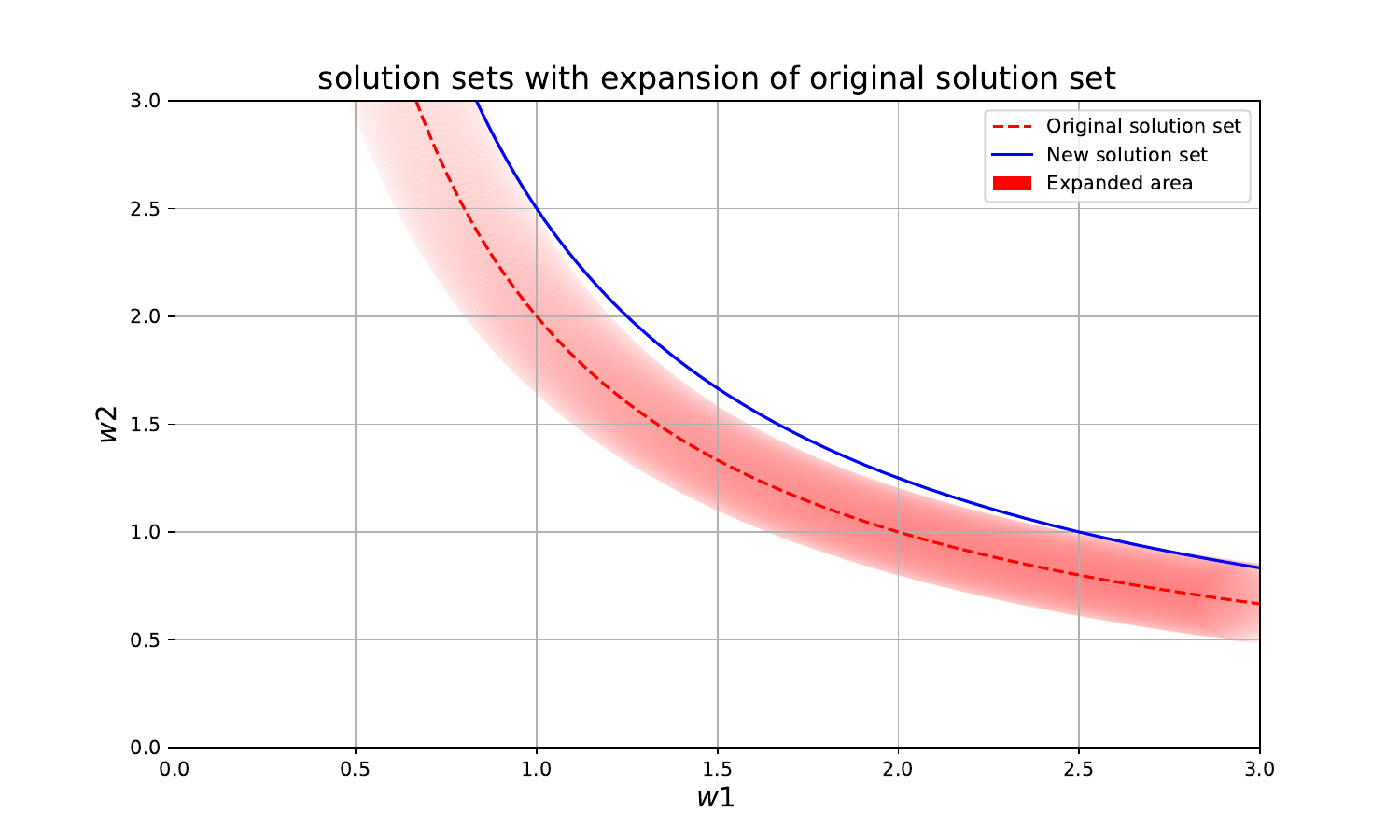}
    \caption{The location of the original solution set $S(\bar{x})$ and the new solution set $S(x^p)$. It presents how the $S(\bar{x})$ expands to achieve $S(x^p)$, where the red area is the expanded area of $S(\bar{x})$ as $S(\bar{x})+\kappa\left\|x-x^p\right\| \mathbb{B}$.}
    \label{fig:theorem}
\end{figure}

Following Lemma \ref{lemma:lip} (see appendix \ref{appendix B}) and Theorem \ref{themrem:1}, the Lipschitz module $\kappa$ for the toy model is 0.2, and $\kappa\left\|x-x^{p}\right\| \mathbb{B}$ now is a unit ball with radius of around 0.178. By the definition in (\ref{def:aubin}), the Lipschitz-like property reveals that the solution set after perturbation, i.e. $S(x^p)$, will fall into the expanded set $S(\bar{x})+\kappa\left\|x-x^{p}\right\| \mathbb{B}$. We plot this fact in Figure \ref{fig:theorem}, which shows that the expanded are of $S(\bar{x})$ is very close to $S(x^p)$.

\begin{figure}[htbp]

    \centering
    \begin{minipage}{0.49\textwidth}
        \includegraphics[width=\linewidth]{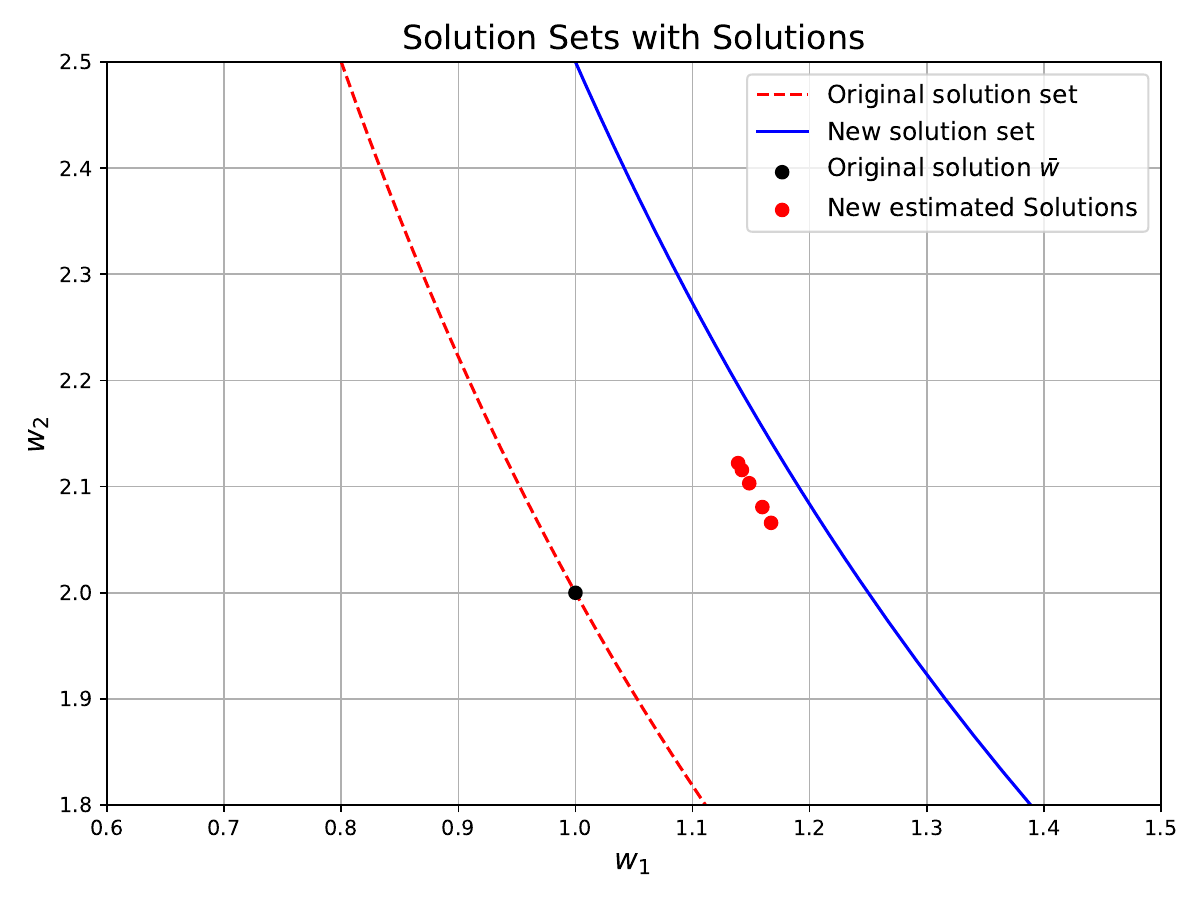}
    \end{minipage}
    \hfill
    \begin{minipage}{0.49\textwidth}
        \includegraphics[width=\linewidth]{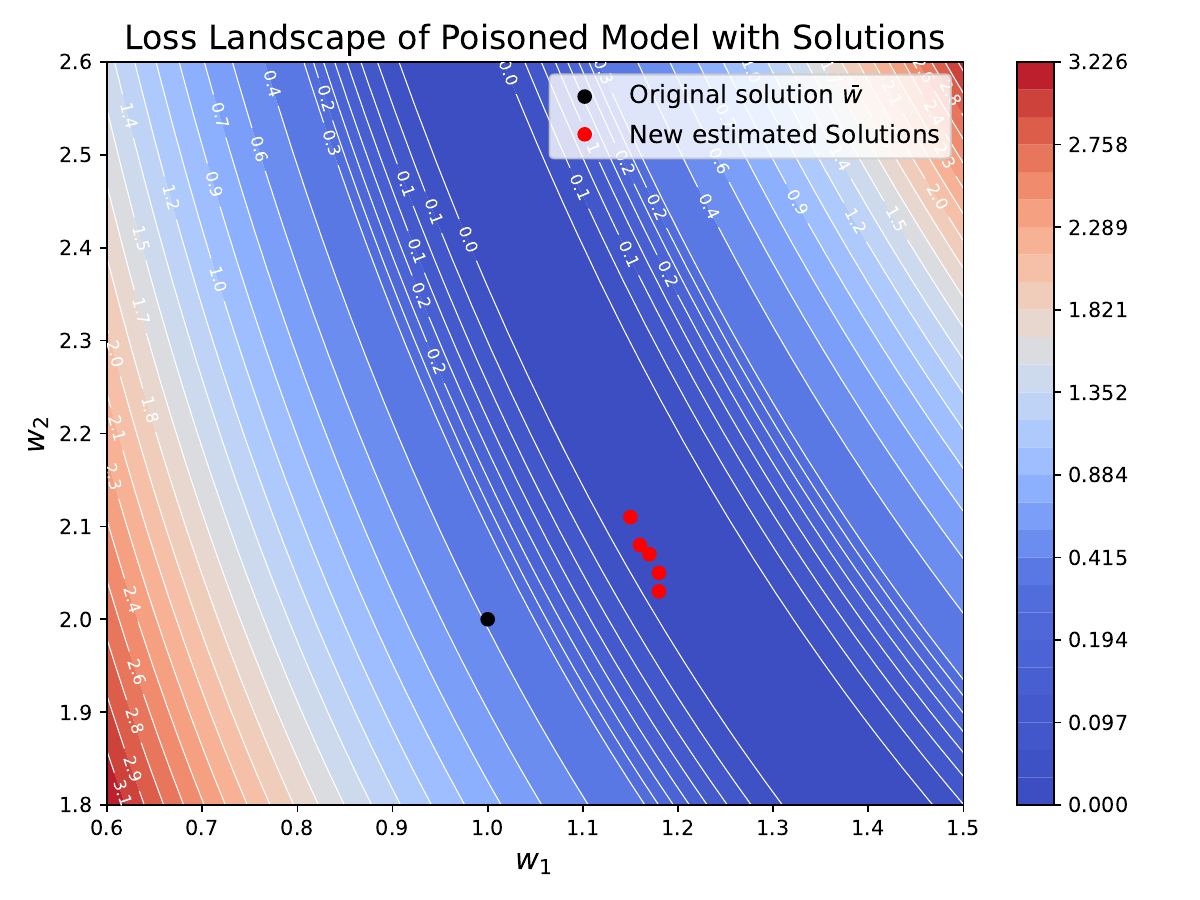}
    \end{minipage}
    \caption{Left: Respective solution set of the pristine and poisoned model. Right: The loss landscape of the poisoned toy model. The black and red points indicate the positions of the original solution $\bar{w}$ and the estimated solutions, respectively. The contour lines represent the corresponding losses.}
    \label{fig:toy model}
\end{figure}

We display five estimated solutions in $S(x^p) \cap \mathbb{B}(\bar{x}, 0.2)$ in Figure \ref{fig:toy model}.
As shown on the left side of Figure \ref{fig:toy model}, both the pristine and poisoned models have their respective set of solutions. The solution set of the poisoned model is shifted upwards compared to the original solution set. Our estimated solutions are very close to the real solution set of the poisoned model. The right side of Figure \ref{fig:toy model} demonstrates the loss landscape of the poisoned model. If we continue to use the original solution $\bar{w}$, the empirical risk would be $0.4$. By shifting the solution from $\bar{w}$ to our estimated solutions, we can decrease the risk to nearly zero (around 0.01).

% The average distance of the estimated solutions in figure \ref{fig:toy model} to $\bar{w}$ is around 0.182. The change in the solution set is close to our estimation. 

% \begin{figure}[htbp]
%     \centering
%     \begin{minipage}[t]{0.45\textwidth}
%         \centering
%         \includegraphics[width=\textwidth]{toy_model_2.pdf}
%         \caption{First Image}
%         \label{fig:image1}
%     \end{minipage}
%     \hfill 
%     \begin{minipage}[t]{0.45\textwidth}
%         \centering
%         \includegraphics[width=\textwidth]{toy_model_2_3d.pdf}
%         \caption{Second Image}
%         \label{fig:image2}
%     \end{minipage}
% \end{figure}

\subsection{Simulation on Resnet}

This section simulates our estimated method, as defined by (\ref{solution estimated method}), on a Resnet56 network \cite{he2016deep}. Following the notations above, we extract 1000 points from the CIFAR-10 dataset, denoted as $\bar{x}$. We use $x^p$ to denote the poisoned data. The original solution $\bar{w}$ is the pre-trained weights of Resnet56.

Following (\ref{solution estimated method}), when the data $\bar{x}$ is perturbed along the direction $\Delta x$, the corresponding change direction of $\bar{w}$, denoted by $\Delta w$, is determined by the graphical derivative $D S(\bar{x} \mid \bar{w})(\Delta x)$. The relationship between $\Delta x$ and $\Delta w$ is:
\begin{equation}
\nabla_w^2 \frac{1}{|I|} \sum_{i \in I} L\left(x_i, y_i, w\right) \Delta w+\nabla_x \nabla_w \frac{1}{|K|} \sum_{i \in K} L\left(x_i, y_i, w\right) \Delta x=0.
\end{equation}

Denoting $\dagger$ the pseudo inverse operator, $\Delta w$ is given by:

\begin{equation}\label{eq:equation for deltaw}
\Delta w=-\left(\nabla_w^2 \frac{1}{|I|} \sum_{i \in I} L\left(x_i, y_i, w\right)\right)^{\dagger}\left(\nabla_x \nabla_w \frac{1}{|K|} \sum_{i \in K} L\left(x_i, y_i, w\right) \Delta x\right) .
\end{equation}

Note (\ref{eq:equation for deltaw}) is very similar to (\ref{Dini}) derived from the Dini implicit function theorem. However, different from (\ref{Dini}), (\ref{eq:equation for deltaw}) here does not rely on a strong convex loss function $L$. The key technical challenge lies in solving (\ref{eq:equation for deltaw}) under high-dimensional cases. We transform 
(\ref{eq:equation for deltaw}) to a least square problem:
\begin{equation}\label{LSE}
\Delta w := \underset{v}{\operatorname{argmin}}\frac{1}{2}\left\| \nabla_w^2 \frac{1}{|I|} \sum_{i \in I} L\left(x_i, y_i, w \right)v +\nabla_x \nabla_w \frac{1}{K} \sum_{i \in K} L\left(x_i, y_i, w\right) \Delta x\right\|^2,
\end{equation}

where both $\nabla_x \nabla_w  L\left(x_i, y_i, w\right) \Delta x$ and 
$\nabla_w^2 L\left(x_i, y_i, w\right) v$ can be calculated using implicit Hessian-vector products (HVP) \cite{pearlmutter1994fast}. Following \cite{agarwal2017second}, $\nabla_w^2 L\left(x_i, y_i, w\right) v$ can be computed efficiently in $O(p)$.

% \begin{table}[ht]
% \centering
% \caption{Empirical loss of the Resnet56 using original and estimated solutions}
% \label{tab:loss_resnet}
% \begin{tabular}{ccc}
% \toprule
% Solution & \begin{tabular}[c]{@{}c@{}}Average Loss on \\the whole corrupted dataset\end{tabular} & \begin{tabular}[c]{@{}c@{}}Average Loss on\\ the perturbed data points\end{tabular} \\ 
% \midrule
% Original Solution & 0.1832 & 0.7257 \\
% Estimated Solution 1 & 0.0238 & 0.0343 \\
% Estimated Solution 2 & 0.0347 & 0.0432 \\
% Estimated Solution 3 & 0.0275 & 0.0433 \\
% Estimated Solution 4 & 0.0299 & 0.0404 \\
% \bottomrule
% \end{tabular}
% \end{table}

% \begin{table}[ht]

% \centering
% \caption{Empirical loss of the Resnet56 on the perturbed points using original and estimated solutions}
% \label{tab:1}
% \begin{tabular}{cccc}
% \toprule
% \diagbox{Solution}{perturbation} & \begin{tabular}[c]{@{}c@{}}Perturb one point\end{tabular} & \begin{tabular}[c]{@{}c@{}}Perturb 10 points\end{tabular} & \begin{tabular}[c]{@{}c@{}}Perturb 500 point\end{tabular} \\ 
% \midrule
% Original Solution &0.0041& 0.0045 & 0.0131  \\
% Estimated Solution &\num{1.9e-5}& \num{6.2e-5} & 0.0027 \\
% \bottomrule
% \end{tabular}
% \end{table}

\begin{table}[ht]

\centering
\caption{Empirical loss of the Resnet56 on the perturbed points using original and estimated solutions}
\label{tab:1}
\begin{tabular}{cccc}
\toprule
\diagbox{Solution}{perturbation} & \begin{tabular}[c]{@{}c@{}}Perturb one point\end{tabular} & \begin{tabular}[c]{@{}c@{}}Perturb 10 points\end{tabular} \\ 
\midrule
Original Solution &0.0041& 0.0045   \\
Estimated Solution &\num{1.9e-5}& \num{6.2e-5} \\
\bottomrule
\end{tabular}
\end{table}

% \begin{table}[ht]
% \centering
% \caption{Empirical loss of the Resnet56 using original and estimated solutions}
% \label{tab:loss_resnet}
% \begin{tabular}{cccc}
% \toprule
% Solution & \begin{tabular}[c]{@{}c@{}}Average Loss on the \\ whole corrupted dataset\end{tabular} & \begin{tabular}[c]{@{}c@{}}Average Loss on the \\ 250 perturbed data points\end{tabular} & \begin{tabular}[c]{@{}c@{}}Average Loss on the \\ whole pristine dataset\end{tabular} \\ 
% \midrule
% Original Solution & 0.0016 & 0.0131 & 0.0015 \\
% Estimated Solution & 0.0025 & 0.0027 & -- \\
% % Estimated Solution 2 & 0.0008 & 0.0011 & -- \\
% % Estimated Solution 3 & 0.0275 & 0.0433 & -- \\
% % Estimated Solution 4 & 0.0299 & 0.0404 & -- \\
% \bottomrule
% \end{tabular}
% \end{table}

% Table \ref{tab:loss_resnet} reports the average loss of the Resnet56 on the entire corrupted dataset and on only the perturbed data points, comparing the loss when the model continues to use the original solution versus adopting our estimated solutions. 

\textbf{Comparison of Loss Between Original and Estimated Solutions on Perturbed Data} We perturb an individual point $x_k$ along the direction of $\nabla_{x_k} L(\bar{w})$. If we continue to use the $\bar{w}$ as our weights, the training loss for the points perturbed would increase from near zero to around 0.0041, due to the perturbations. This indicates that the original solution is far from the solution set of the poisoned model.  We estimate the change of $\bar{w}$, i.e. $\Delta w$ through solving the problem (\ref{LSE}). By moving the original solution along $\Delta w$, we obtain the estimated solution. By adopting our estimated solutions, the training loss decreases to near zero, demonstrating that our estimated solution change $\Delta w$ effectively draws $\bar{w}$ towards the solution set of the poisoned model. Table \ref{tab:1} reports the loss on the perturbed points for varying numbers of perturbed points when we adopt the original solution and the estimated solution. 

\begin{figure}[htbp]

    \centering
    \begin{minipage}{0.49\textwidth}
        \includegraphics[width=\linewidth]{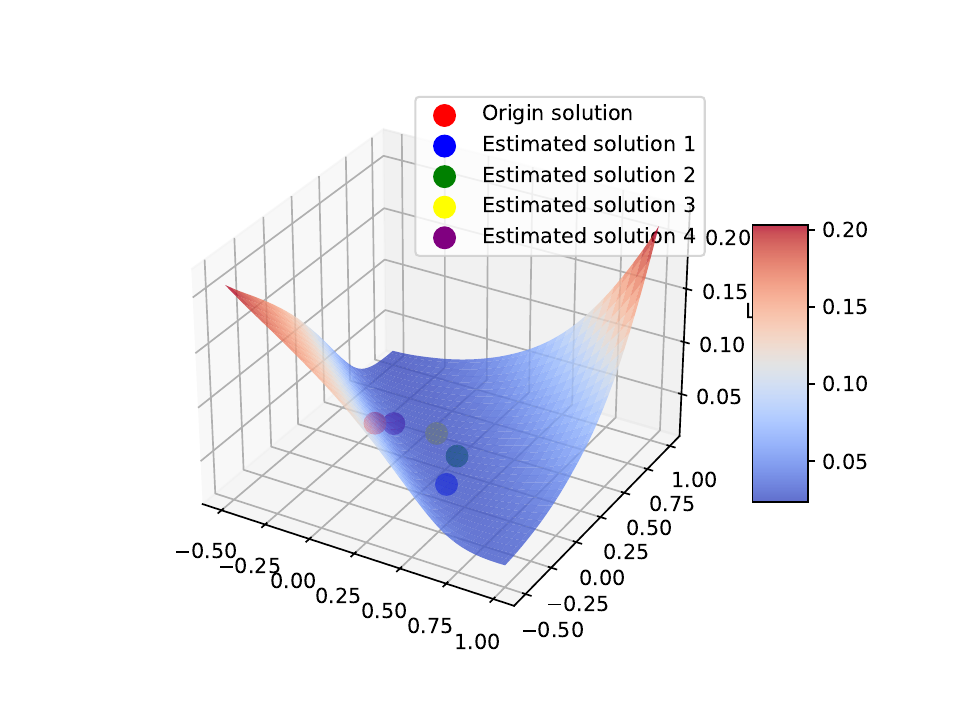}
    \end{minipage}
    \hfill
    \begin{minipage}{0.49\textwidth}
        \includegraphics[width=\linewidth]{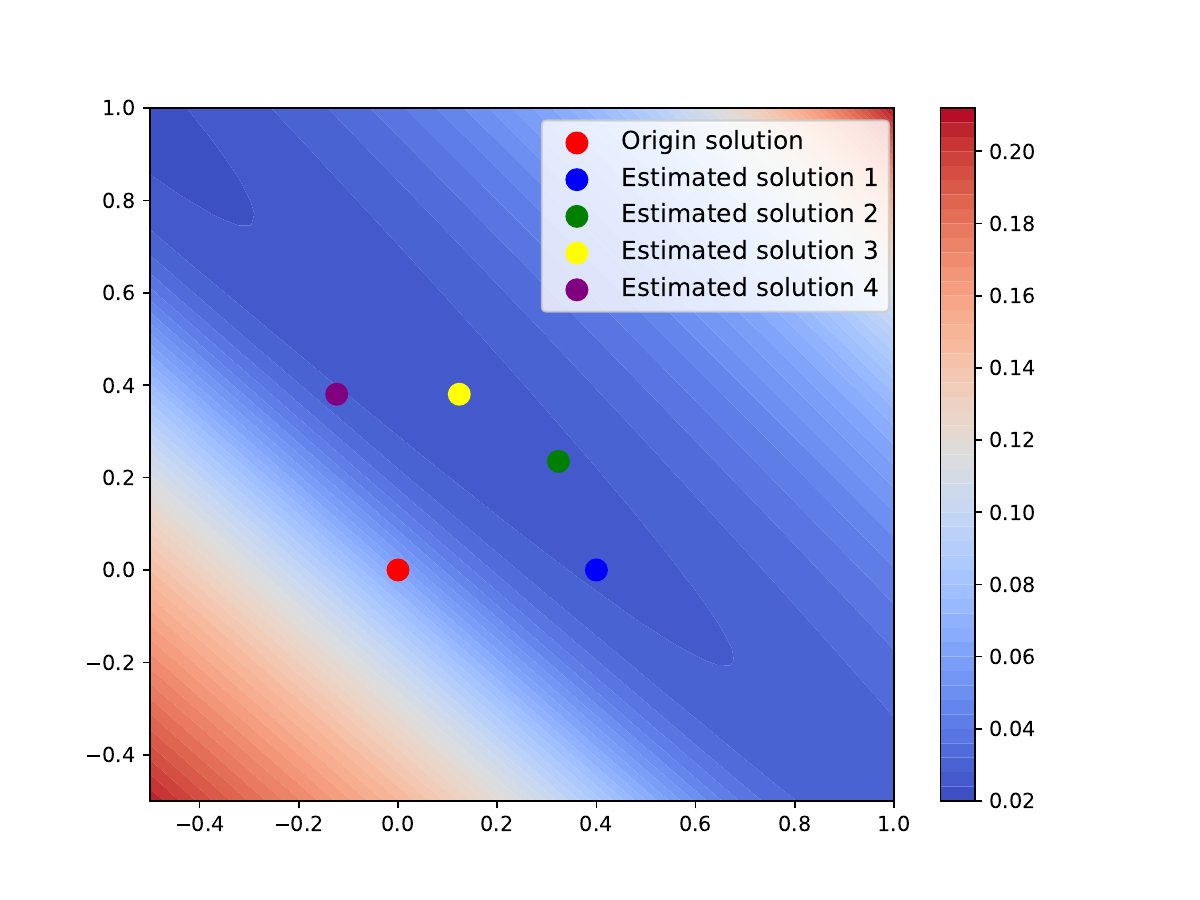}
    \end{minipage}
    \caption{3D image(left) and  2D contours(right) of the loss landscape of poisoned Resnet56, indicating how the $\Delta w$ draws the original solution towards the valley of the landscape. }
    \label{fig:resnet loss}
\end{figure}

We then perturb 500 points and run problem (\ref{LSE}) multiple times, deriving different numerical results for $\Delta w$. Figure \ref{fig:resnet loss} demonstrates the location of the original solution and estimated solutions on the loss landscape of poisoned Resnet56.  The loss landscape visualization follows the work in \cite{li2018visualizing}, where the color intensity and the left plot's vertical axis indicate the loss. By moving the original solution $\bar{w}$ along $\Delta w$, the solutions reach the valley of the landscape of the poisoned model. Each estimated solution in Figure \ref{fig:resnet loss} corresponds to a simulation result of (\ref{LSE}).

% Table \ref{tab:loss_resnet} reports the average loss of the Resnet56 on the entire corrupted dataset and on only the perturbed data points, comparing the loss when the model continues to use the original solution versus adopting our estimated solutions.

\textbf{Comparison of Actual and Estimated Solution Changes} 
Under the 10 data perturbation scenarios, we calculate the actual difference in the model solution (e.g., parameters) before and after data perturbation and compare it with the estimated solution change derived from Algorithm \ref{alg}.  The difference is quantified using the relative difference (relative difference between $w_1$ and $w_2$ is defined as $\frac{\left|w_1-w_2\right|}{\left|w_1\right|}$). The results are summarized in Table~\ref{table2}.

\begin{table}[h!]
\caption{Relative differences between solutions under perturbation scenarios.}
\label{table2}
\centering
\renewcommand{\arraystretch}{1.2} % Adjust row spacing
\begin{tabular}{l c}
\toprule
\textbf{Comparison} & \textbf{Relative Difference} \\
\midrule
Original vs. Retrained Solution & \( 5.37 \times 10^{-5} \) \\
Original vs. Estimated Solution & \( 19 \times 10^{-5} \) \\
\bottomrule
\end{tabular}

\label{tab:relative_diff}
\end{table}

\textbf{Tolerance to Perturbations with Varying Magnitudes}  
We perturbed the data point \(x\) by adding \(\text{stepsize} \cdot \|x\| \cdot \text{sign}(\nabla_x \text{Loss})\). The results below demonstrate the method's tolerance to perturbations with stepsize magnitudes up to 2\% of the norm of the data.

\begin{table}[h!]
\centering
\caption{Empirical loss of ResNet56 on the perturbed points using the original and estimated solutions.}
\renewcommand{\arraystretch}{1.2} % Adjust row spacing
\begin{tabular}{c c c}
\toprule
\textbf{Stepsize (\%)} & \textbf{Original Solution} & \textbf{Estimated Solution} \\
\midrule
0.20 & 0.0045 & \(6.20 \times 10^{-5}\) \\
0.50 & 0.15   & 0.005 \\
1    & 1.19   & 0.3 \\
2    & 3.35   & 1.9 \\
\bottomrule
\end{tabular}

\label{tab:resnet56_loss}
\end{table}

\section{Related work}

This paper only focuses on the sensitivity of solutions of learning models (e.g., model weights of DNNs) in response to perturbations in the training data. For the change of prediction in relationship to the inference data, one can refer to \cite{fazlyab2019efficient,weng2018evaluating}. Influence function, as a concept in robust statistics \cite{law1986robust}, was first used for the sensitivity analysis of the classification models with convex loss function (e.g. SVM, Logistic Regression) \cite{christmann2004robustness}. Koh et al. \cite{koh2017understanding} introduced it to the DNN, demonstrating its application in data poisoning attacks and identifying the important features. However, the existence of the influence function relies on the implicit function theorem (see Theorem 19 in \cite{christmann2004robustness}), which may not be applicable to DNNs when non-isolated DNN solutions are considered, as discussed in the Introduction section of this paper. Peter et al. \cite{nickl2024memory} measured the sensitivity of solutions to training data through the Memory-Perturbation Equation (MPE). It was demonstrated that sensitivity to a group of examples can be estimated by adding their natural gradients, indicating higher sensitivity with larger gradients. However, its theorem relies on the inverse of Hessian, which does not exist when the loss function is not strongly convex.

This paper utilizes the Lipschitz module to characterize the sensitivity of DNN. Noteworthy is that sensitivity analysis in our paper studies how the model \textit{solution} changes with training data, not how the model \textit{output} changes with inference data, although the two are closely related. To our best knowledge, this is the first time that the Lipschitz concept and the estimation of Lipschitz module estimation are introduced to the training stage. Previous research only focused on estimating the Lipschitz constants during the inference stage, see \cite{fazlyab2019efficient,virmaux2018lipschitz}, to quantify the robustness of model prediction w.r.t. perturbations in inference data. For example, Kevin and Aladin \cite{virmaux2018lipschitz} adopted a power method working with auto differentiation to estimate the upper bound of the Lipschitz constant. 

Most sensitivity analysis approaches only focus on a single solution of the learning algorithm. This paper considers the first-order optimality condition as a set-valued mapping (multifunction), introducing the 'set-to-set' analysis approach. Our sensitivity analysis is based on the Lipschitz-like property of set-valued mapping, where the Lipschitz Modulus quantifies the change of the solution set. For more discussion about set-valued mapping, interested readers can refer to \cite{rockafellar2009variational,dontchev2009implicit}.

\section{Conclusion}

This paper provides set-valued analysis methods to study the sensitivity of model solutions (e.g. weights of a DNN) in response to perturbations in the training data. Theoretically, our approach considers the possibility that the DNN may not have unique solutions and does not rely on a non-singular Hessian. We also accurately estimate the solution change when the training data are perturbed along a specific direction.

Our analysis framework can have multiple potential applications. First, it utilizes the Lipschitz concept 
to study the sensitivity of DNN. This can lead to a robustness evaluation method by measuring the Lipschitz module, i.e. a larger Lipschitz module indicates higher sensitivity. Second, our framework extends the implicit function theorem in DNN, which can be used for data poisoning attacks. We can determine the perturbation direction of training data to shift the solution toward a target solution, thereby executing a model target poisoning attack \cite{suya2021model}. Alternatively, we can identify the perturbation direction to alter the training data to increase the loss of validation data \cite{mei2015using}.

In future research, we plan to test the set-valued analysis methods on more DNNs. One limitation of this research is that our theoretical results are derived only using the DFCNN with the Relu activation function. In the future, we plan to extend the results to a wider variety of DNN architectures.

\newpage

\small \bibliographystyle{plain}
\bibliography{ref} %

\newpage 
\appendix
\section{Appendix A} \label{appendix A}
This section provides the proof of the Theorem (\ref{themrem:1}). We begin by restating two lemmas to support our proof.

\begin{lemma}(see Theorem 6.14 and Theorem 6.31in \cite{rockafellar2009variational})\label{lemma3}

Let
\begin{equation*}
    C=\{x \in X \mid G(x) \in D\},
\end{equation*}
for closed sets $X \subset \mathbb{R}^d$ and $D \subset \mathbb{R}^p$ and a $\mathcal{C}^1$ mapping $F: \mathbb{R}^d \rightarrow \mathbb{R}^p$, written componentwise as $G(x)=\left(g_1(x), \ldots, g_m(x)\right)$. At any $\bar{x} \in C$ satisfying the constraint qualification that
\begin{equation}\label{18}
\left\{\begin{array}{l}
\text { the only vector } y \in N_D(G(\bar{x})) \text { for which } \\
-\sum_{i=1}^m y_i \nabla g_i(\bar{x}) \in N_X(\bar{x}) \text { is } y=(0, \ldots, 0) .
\end{array}\right.
\end{equation}

If in addition to this constraint qualification the set $X$ is regular at $\bar{x}$ and $D$ is regular at $F(\bar{x})$, then $C$ is regular at $\bar{x}$ and
\begin{equation}
N_C(\bar{x})=\left\{\sum_{i=1}^m y_i \nabla g_i(\bar{x})+z \mid y \in N_D(G(\bar{x})), z \in N_X(\bar{x})\right\}
\end{equation}

\begin{equation}
T_C(\bar{x})=\left\{w \in T_X(\bar{x}) \mid \nabla G(\bar{x}) w \in T_D(G(\bar{x}))\right\}
\end{equation}
\end{lemma}

\textit{proof of Theorem \ref{themrem:1}}

For the mapping $F\left(x_k\right)=\left\{w \left\lvert\, \nabla_w \frac{1}{n} \sum_{i=1}^n L\left(x_i, y_i, w\right)=0\right.\right\}=\left\{w \mid R(x, w)=0\right\}$,we have,

\begin{equation}\label{gpfF}
  \operatorname{gph} F = \left\{(x_k,w) \in  \mathbb{R}^{d+p}\mid R(x_k,w)=0\right\}
\end{equation}

We consider $\nabla_w \frac{1}{n} \sum_{i=1}^n L\left(x_i, y_i, w\right)=0$ as a multifunction with \begin{equation}
    (\nabla_{w_1} \frac{1}{n} \sum_{i=1}^n L\left(x_i, y_i, w\right),\dots,\nabla_{w_p} \frac{1}{n} \sum_{i=1}^n L\left(x_i, y_i, w\right)=\mathbf{0}\in \mathbb{R}^p
\end{equation}

We use $R_i(x,w)$ to denote $\nabla_{w_i} \frac{1}{n} \sum_{i=1}^n L\left(x_i, y_i, w\right)$.

Let $X=\mathbb{R}^{d+p}$ ,$ D=\left\{0 \right\}$.
Following the definition \ref{def3} and definition \ref{def4},we obtain 
$N_D(R(\bar{x}_k,\bar{w}))=\mathbb{R}^p$,$\left.T_D(R(\bar{x},\bar{w}))\right\}=0$, $T_X(\bar{x}_k)=\mathbb{R}^{d+p}$ and $N_x(R(\bar{x},\bar{w}))=0$. Under the assumption 2, for any vector $y \in N_D(R(\bar{x}_k,\bar{w})=\mathbb{R}^p$, if $\sum_{i=1}^p y_i\left[\nabla_{x_k} R_i(\bar{x}, \bar{w}) ; \nabla_w R_i(\bar{x}, \bar{w})\right]=0$, $y$ is a zero vector because $y_i\left[\nabla_w R(\bar{x}, \bar{w}), \nabla_{x_k} R(\bar{x}, \bar{w})\right]$ is of full rank. $(\bar{x}_k,\bar{w}) \in \operatorname{gph} F$ satisfies the constraint qualification (\ref{18}).

Following the definition \ref{def:genedev},
\begin{equation}
D F\left(\bar{x}_k \mid \bar{w}\right)(\mu)=\left\{v \mid (\mu, v) \in T_{\operatorname{gph} F}(\bar{x}_k, \bar{w})\right\},
\end{equation}
\begin{equation}
D^* F\left(\bar{x}_k \mid \bar{w}\right)(p)={q \mid (q,-p) \in N_{\operatorname{gph} F}(\bar{x}, \bar{w})}.
\end{equation}

Denoting $X=\mathbb{R}^{d+p}$ ,$ D=\left\{0 \right\}$, apply the lemma \ref{lemma3} to (\ref{gpfF}), we have:
\begin{equation}\label{25}
    T_{\operatorname{gph} F}(\bar{x}_k, \bar{w})=\left\{(\mu,v)\in T_X\left(\bar{x}_k,\bar{w}\right)\mid [\nabla_x R(\bar{x},\bar{w});\nabla_w R(\bar{x},\bar{w})](\mu,v)^T \in T_D(R(\bar{x},\bar{w}))\right\}
\end{equation}
\begin{equation}\label{26}
    N_{\operatorname{gph} F}(\bar{x}_k, \bar{w})=\left\{\sum_{i=1}^p y_i [\nabla_{x_k}R_i(\bar{x},\bar{w});\nabla_{w}R_i(\bar{x},\bar{w})]+z \mid y \in N_D(R(\bar{x},\bar{w})), z \in N_X(\bar{x}_k)\right\}
\end{equation}

Substituting $N_D(R(\bar{x}_k,\bar{w}))=\mathbb{R}^p$,$\left.T_D(R(\bar{x},\bar{w}))\right\}=0$, $T_X(\bar{x}_k)=\mathbb{R}^{d+p}$ and $N_x(R(\bar{x},\bar{w}))=0$ to (\ref{25}) and (\ref{26}) lead to theorem \ref{themrem:1}.

\textbf{Remark:} Similiar to the proof of theorem \ref{themrem:1}, applying lemma \ref{lemma3} to the mapping $F(x)=\left\{w \left\lvert\, R(x, w)=\nabla_w \frac{1}{n} \sum^n L\left(x_i, y_i, w\right)=0\right.\right\}$  leads to (\ref{eq:solution estimation}).

\section{Appendix B} \label{appendix B}

\textbf{Notation:} Given the dataset $x_i \in \mathbb{R}^d, i \in I=[1,2,\dots,n]$, assume we only perturb a single data $x_k, k \in I$, where $ x_k=\left\{x_{p}^1, \ldots, x_{p}^d\right\}$, $x_{p}^m \in R$ is the $m$-th component of $x_k$.

% \textbf{Lemma A.1 }For any $m \in[1,2 \ldots d ]$, the mapping
% $$
% F_h^m\left(x_{\text {p}}^m\right)=\left\{w^{(h)} \mid R(\bar{x}, \bar{w})=0\right\}
% $$
% holds Lipschitz-Lile property.
We first introduce two lemmas utilizing a mapping $G$ to support our proof.

\begin{lemma} (see \cite{dontchev1996characterizations}) \label{lamma:aubin}
 If a set-valued mapping $G: X \Rightarrow W$ is locally closed at $(\bar{x}, \bar{w}) \in \operatorname{gph} G$, then $G$ has the Aubin property at $\bar{x}$ for $\bar{w}$ if and only if
\begin{equation}
D^* G(\bar{x}\mid \bar{w})(0)=\{0\} .
\end{equation}   
\end{lemma}

\begin{lemma}\label{lemma:lip}(see Theorem4.37 in \cite{mordukhovich2006variational})
Let
\begin{equation}
G(x):=\{w \in W \mid R(x, w)\in \Theta\},
\end{equation}
where $R: X \times W \rightarrow Z$ with $R(\bar{x}, \bar{w})=\bar{z}$.
Assume  $G$ is strictly differentiable at $(\bar{x}, \bar{w})$ with the surjective derivative $\nabla R(\bar{x}, \bar{w})$, $\Theta$ is locally closed around $\bar{z}$. Then the Lipchitz modulus  of $G$ at $(\bar{x}, \bar{w})$ satisfies:

\begin{equation}\label{eq:lip}
\operatorname{lip} G(\bar{x}, \bar{w})=\sup \left\{\left\|x^*\right\| \mid\left(x^*,-w^*\right) \in \nabla R(\bar{x}, \bar{w})^* N(\bar{z} ; \Theta),\left\|w^*\right\| \leq 1\right\}.
\end{equation}
\end{lemma}

\textbf{\textit{proof of Theorem2}}
Under the Assumption 1, $R(\bar{x}, \bar{w})=0$. This implies that
\begin{equation}
\left(f\left(\bar{x}_i, \bar{w}\right)-\bar{y}_i\right)^2 / 2=0, \forall i \in I.
\end{equation}

By the first-order optimality condition, we have
\begin{equation}
\left(f\left(\bar{x}_i, \bar{w}\right)-\bar{y}_i\right) \nabla_w f\left(\bar{x}_i, \bar{w}\right)=0.
\end{equation}
For each $i \in I$, we define mapping $S_i$:
\begin{equation}\label{eq:s_i}
S_i\left(x\right)=\left\{w \mid\left(f\left(x_i, w\right)-y_i\right) \nabla_w f\left(x_i, w\right)=0\right\},
\end{equation}
where $S_i\left(x\right)$ is the solution set such that $\left(f\left(\bar{x}_i, \bar{w}\right)-\bar{y}_i\right)^2 / 2=0$.

Given $\bar{x}=\left[\bar{x}_1, \bar{x}_2, \ldots, \bar{x}_n\right] $ and $\bar{w}^{(h)}$, for $i \ne k$, since $x_i$ will not be perturbed, and the perturbation on $x_k$ has no impact
on solution sets $S_i(x)$, we have,

\begin{equation}\label{eq:s1=s2}
S_i\left(x\right)=S_i\left(x^{\prime}\right), \forall x, x^{\prime} \in V, i \ne k.
\end{equation}

For $i=k$, we now prove that mapping $S_k$ holds Lipschitz-like property

For fully connected NN defined in section \ref{sec:MLP}, we have
\begin{equation}
\frac{\partial f(x_i, w)}{\partial W^{(h)}}=\operatorname{diag}(\mathbf{1}(\sigma(W^{(h)} x_i^{h-1})>0))\left[\prod_{k=h+1}^H {W^{(k)}}^{\top} \operatorname{diag}(\mathbf{1}(\sigma(W^{(k)} x_i^{k-1})>0))\right] a {x_i^{(h-1)}}^{\top},
\end{equation}

\begin{equation}
\mathbf{R}^d \ni \frac{\partial f(x_i, w)}{\partial x^{(k)}}=\left[\prod_{k=1}^H {W^{(k)}}^{\top} \operatorname{diag}(\mathbf{1}(\sigma(W^{(k)} x_i^{k-1})>0))\right] a,
\end{equation}
We set$\frac{\partial f(x_i, w)}{\partial w^{(h)}}$ to be the flatten vector of $\frac{\partial f(x_i, w)}{\partial W^{(h)}}$, $\frac{\partial f(x_i, w)}{\partial w}=[\frac{\partial f(x_i, w)}{\partial w^{(1)}};\frac{\partial f(x_i, w)}{\partial w^{(2)}};\dots, \frac{\partial f(x_i, w)}{\partial w^{(H)}}]^T$.

Following the definition \ref{def:derivative},  let $p:=\Delta w^{(h)} \in \mathbb{R}^{dim(w^{(h)})}$,$q:=\Delta x_k \in \mathbb{R}^d$ \\, $R_k:=\left(f\left(x_k, w\right)-y_i\right) \nabla_w f\left(x_k, w\right)$. We can derive the coderivative of $S_k^*$ using theorem \ref{themrem:1}

\begin{equation}
S_k^*\left(x_k \mid w^{(h)}\right)\left(\Delta w^{(h)}\right)=\left\{\Delta x_k \mid\left[\Delta x_k,-\Delta w^{(h)}\right]=\left[\nabla x_k R_k, \nabla_{w^{(h)}} R_k\right]^{\top} y, y \in \mathbb{R}^{dim(w^{})}\right\}.
\end{equation}

where 
\begin{equation}
\label{eq:1}
\nabla_{w^{(h)}} R_k=\frac{\partial f\left(x, w\right)}{\partial w}\left(\frac{\partial f(x,w)}{\partial w^{(h)}}\right)^{\top} \in \mathbb{R}^{p \times p^{(h)}}, p=dim (w), p^{(h)}=dim\left(w^{(h)}\right), 
\end{equation}

\begin{equation}\label{eq:3}
\nabla x_k R_k=\frac{\partial f\left(x, w\right)}{\partial w}\left(\frac{\partial f(x,w)}{\partial x_{k}}\right)^{\top} \in \mathbb{R}^{p\times d}
\end{equation}

For the given $\bar{x}_k$ and $\bar{w}$, if $\Delta \bar{w}^{(h)}=0$, following equation (\ref{eq:1}) ,we have $\left(\frac{\partial f(\bar{x}_k,\bar{w})}{\partial w^h}\right)_i\left(\frac{\partial f(\bar{x}_k,\bar{w})}{\partial w}\right)^{\top} y=0,$, where$\left(\frac{\partial f\left(\bar{x}_k, \bar{w}\right)}{\partial w^h}\right)_i$ is $i$-th component of $\frac{\partial f\left(\bar{x}_k, \bar{w}\right)}{\partial w^h}$, $1 \leqslant i \leqslant p^{(h)}$. Thus, we have 
\begin{equation}\label{eq:2}
\left(\frac{\partial f\left(\bar{x}_k, \bar{w}\right)}{\partial w}\right)^{\top} y=0.
\end{equation}

Combining equation (\ref{eq:2}) and equation (\ref{eq:3}) implies that $\Delta x_k=0$. Following 
lemma \ref{lamma:aubin}, mapping $S_k$ holds the Lipschitz-like property, i.e. there exists neighborhoods $V_k$ of $\bar{x}_k$ and $U_h$ of $\bar{w}^{(h)}$, with a positive real number $\kappa_h$ (called Lipschitz modulus) such that
\begin{equation}
S_k\left(x_k^{\prime}\right) \cap U_h \subset S_h\left(x_k\right)+\kappa_h\left\|x_k-x_k^{\prime}\right\| \mathbb{B} \quad \forall x_k^{\prime}, x_k \in V_k
\end{equation}

Observe that each $L\left(x_i, y_i, w\right)$ attains a value of zero when $x=\bar{x}$ and $w=\bar{w}$. Consequently, the parameter $F_h(x_k)$ of the fully connected neural network can be interpreted as the intersection of the solution sets $S_i(x_k)$. Each set $S_i(x_k)$ corresponds to the solutions for $w$ such that $L(x_i,y_i,w)=0$."
\begin{equation}\label{eq:cap}
 F_h(x_k)=S_1(x_k) \cap S_2(x_k),\dots, \cap S_k(x_k)\dots, \cap S_n(x_k) 
\end{equation}

Substitute (\ref{eq:s1=s2}) into (\ref{eq:cap}), we have:

\begin{equation}
\begin{aligned}
   F_h(x_k^\prime)\cap U_h -F_h(x_k)&\subset F_h(x_k^\prime)-F_h(x_k)\\
   &=S_1(x_k^\prime)\cap\dots\cap S_k(x_k^\prime)\dots \cap S_n(x_k^\prime)- S_1(x_k)\cap\dots \cap S_k(x_k)\dots, \cap S_n(x_k) \\
   &=S_1(x_k)\cap\dots, \cap S_k(x_k^\prime)\dots, \cap S_n(x_k)- S_1(x_k)\cap\dots \cap S_k(x_k)\dots, \cap S_n(x_k) \\  
   & =S_1(x_k) \cap S_2(x_k)\cap\dots \cap (S_k(x_k^\prime)-S_k(x_k))\dots \cap S_n(x_k)\\
   &\subset \kappa_h\left\|x_k-x_k^{\prime}\right\| \mathbb{B}
\end{aligned}
\end{equation}

By applying Lemma \ref{lemma:lip} to the mapping $S_k$, and substituting $\Theta=\{0\}$, $\bar{z}=0$ and $N(\bar{z} ; \Theta)=\mathbf{R}^p$, we obtain:
\begin{equation}
\begin{aligned}
\kappa_h &= \| \frac{\partial f\left(x_k, w\right)}{\partial x^{(k)}} \| / \| \frac{\partial f\left(x_k, w\right)}{\partial w^{(h)}}\| \\
&= \frac{\| \left[\prod_{k=1}^H W^{(k)^{\top}} \operatorname{diag}\left(\mathbf{1}\left(\sigma\left(W^{(k)} x_k^{k-1}\right)>0\right)\right)\right] a\|}{\| \operatorname{diag}\left(\mathbf{1}\left(\sigma\left(W^{(h)} x_k^{h-1}\right)>0\right)\right)\left\lfloor\prod_{k=h+1}^H W^{(k)^{\top}} \operatorname{diag}\left(\mathbf{1}\left(\sigma\left(W^{(k)} x_k^{k-1}\right)>0\right)\right)\right] a x_k^{(h-1)^{\top}} \|_F}
\end{aligned}
\end{equation}

\textit{\textbf{proof of Theorem \ref{thm3}:}}
Assume
\begin{equation}
x^{\prime}=\left[x_1^{\prime}, x_2^{\prime} \ldots x_n^{\prime}\right]=\left[x_1+\Delta x_1, x_2+\Delta x_2 \ldots x_n+\Delta x_n\right]
\end{equation}

% $$
% F_n(X
% $$
\begin{equation}\label{28}
\begin{aligned}
& F_h\left(x^{\prime}\right)-F_n(x) \\
= & F_h\left(x_1+\Delta x_1, x_2+\Delta x_2, \ldots, x_{n-1}+\Delta x_{n-1},x_n+\Delta x_n\right)-F_h\left(x_1+\Delta x_1, x_2+\Delta x_2, \ldots, x_{n-1}+\Delta x_{n-1},x_n )\right. \\
+ & F_h\left(x_1+\Delta x_1, x_2+ \Delta x_2, \ldots, x_{n-1}+\Delta x_{n-1},x_n\right)-F_h\left(x_1+\Delta x_1, x_2+\Delta x_2 \ldots x_{n-1}+x_n\right) \\
+ & \ldots \\
+ & F_h\left(x_1+\Delta x_1, x_2, \ldots x_n\right)-F_h\left(x_1, x_2, \ldots x_n\right)
\end{aligned}
\end{equation}

From Theorem \ref{thm:2}, there exists a neighbor $V_i$ of $x_i$, $U_i$ of $\bar{w}$ and $k_i$. st

\begin{equation}\label{29}
\begin{aligned}
&F_h\left(x_1 + \Delta x_1, x_2 + \Delta x_2, \ldots, x_i + \Delta x_i, x_{i+1}, \ldots, x_n\right) \cap U_i \\
&\quad - F_n\left(x_1 + \Delta x_1, x_2 + \Delta x_2, \ldots, x_i, x_{i+1}, \ldots, x_n\right) \\
&\subset \kappa_i \left\| x_i - x_i' \right\| \mathbb{B} \quad \forall x_i', x_i \in V_i
\end{aligned}
\end{equation}

Apply (\ref{29}) to (\ref{28}), we obtain:

\begin{equation}
\begin{aligned}
& F_h\left(x^{\prime}\right)\cap U-F_n(x) \\
\subset & F_h\left(x_1+\Delta x_1, x_2+\Delta x_2, \ldots, x_{n-1}+\Delta x_{n-1},x_n+\Delta x_n\right)\cap U_n-F_h\left(x_1+\Delta x_1, x_2+\Delta x_2, \ldots, x_{n-1}+\Delta x_{n-1},x_n )\right. \\
+ & F_h\left(x_1+\Delta x_1, x_2+ \Delta x_2, \ldots, x_{n-1}+\Delta x_{n-1},x_n\right)\cap U_{n-1}-F_h\left(x_1+\Delta x_1, x_2+\Delta x_2 \ldots x_{n-1}+x_n\right) \\
+ & \ldots \\
+ & F_h\left(x_1+\Delta x_1, x_2, \ldots x_n\right)\cap U_1-F_h\left(x_1, x_2, \ldots x_n\right)\\
\subset & \kappa_1\left\|x_1-x_1^{\prime}\right\| \mathbb{B}+\kappa_2\left\|x_2-x_2^{\prime}\right\| \mathbb{B}+\dots+\kappa_n\left\|x_n-x_n^{\prime}\right\|
\mathbb{B}\\
\subset &\max \left\{\kappa_1, \kappa_2, \ldots, \kappa_n\right\} \sqrt{n}\left\|\mathbf{x}-\mathbf{x}^{\prime}\right\| \mathbb{B}
=\kappa \left\|\mathbf{x}-\mathbf{x}^{\prime}\right\|\mathbb{B} \quad \forall x^{\prime}, x \in V\\
\end{aligned}
\end{equation}
where ${V}=V_1 \times V_2 \times \cdots \times V_n$, ${U}=U_1 \cap U_2 \cap \cdots \cap U_n$, $\kappa=\sqrt{n}\max \left\{\kappa_1, \kappa_2, \ldots, \kappa_n\right\} $
\newpage
\section{Appendix C} \label{appendix C}

\begin{figure}[htbp]

    \centering
    \begin{minipage}{0.49\textwidth}
        \includegraphics[width=\linewidth]{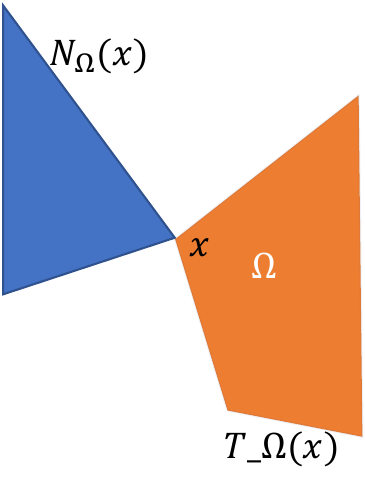}
    \end{minipage}
    \hfill
    \begin{minipage}{0.49\textwidth}
        \includegraphics[width=\linewidth]{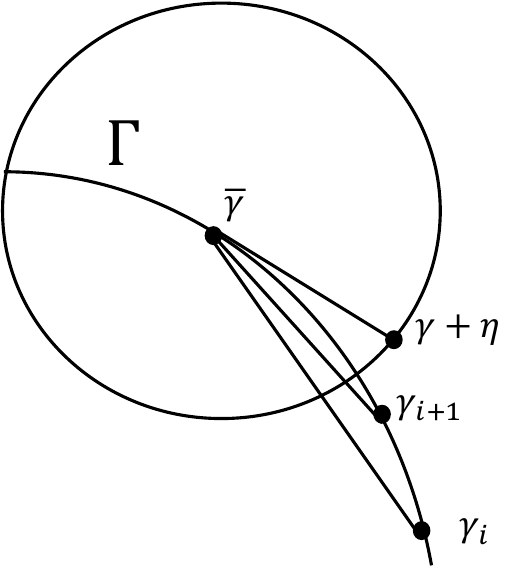}
    \end{minipage}
    \caption{Left: Illustrative diagram showing the normal cone $N_{\Omega}(x)$ and tangent cone $T_{\Omega}(x)$ at point $x$ within the set $\Omega$.. Right: $\eta$ is a tangent vector of $\Gamma$ at $\bar{\gamma}$ if there exists $\left\{\gamma_i\right\} \subset \Gamma$ with $\gamma_i \rightarrow \bar{\gamma}$, and a positive scalar sequence $\tau_i$  such that $\tau_i\rightarrow 0$ with $\left(\gamma_i-\bar{\gamma}\right) / \tau_i \rightarrow \eta$..}
    
\end{figure}

\end{document}